\documentclass[11pt]{article}
\usepackage[letterpaper,margin=1in]{geometry}
\usepackage[T1]{fontenc}
\usepackage{times}
\usepackage[authoryear,round]{natbib}
\setcitestyle{citesep={;},aysep={,},yysep={;}}
\usepackage{amsmath,amssymb,graphicx,booktabs,array,tabularx,xcolor}
\usepackage{microtype}
\usepackage{url}
\usepackage{hyperref}
\definecolor{linkblue}{RGB}{0,82,155}
\hypersetup{colorlinks=true,linkcolor=linkblue,citecolor=linkblue,urlcolor=linkblue,
  pdftitle={ProWorld: Multi-View World Model with Proprioceptive Feedback for Closed-Loop Policy Evaluation},
  pdfauthor={Zijun Wang et al.}}
\definecolor{draftblue}{RGB}{37,86,126}
\definecolor{draftgray}{RGB}{246,248,250}

\newcommand{\SetValue}[2]{\expandafter\def\csname paperValue@#1\endcsname{#2}}
\newcommand{\num}[1]{\ifcsname paperValue@#1\endcsname\csname paperValue@#1\endcsname\else\textcolor{draftblue}{\textbf{XX}}\fi}

\newcommand{\NA}{\textemdash}
\newcommand{\ours}{ProWorld}
\newcommand{\placeholder}[3]{%
  \begingroup\setlength{\fboxsep}{6pt}%
  \fcolorbox{draftblue!55}{draftgray}{\begin{minipage}[c][#1][c]{\dimexpr\linewidth-2\fboxsep-2\fboxrule\relax}
  \centering\small\textbf{#2}\par\medskip
  \begin{minipage}{0.94\linewidth}\centering\footnotesize #3\end{minipage}
  \end{minipage}}\endgroup}
\newcommand{\asset}[4]{\IfFileExists{figures/#1}{\includegraphics[width=\linewidth]{figures/#1}}{\placeholder{#2}{#3}{#4}}}

\newcommand{\E}{\mathbb{E}}

\newcommand{\tablehead}[2][c]{\begin{tabular}[c]{@{}#1@{}}#2\end{tabular}}

\title{\Large\bfseries ProWorld: Multi-View World Model with Proprioceptive Feedback\\for Closed-Loop Policy Evaluation}
\author{%
  Zijun Wang$^{1,\dagger}$\quad
  Tianyi Yan$^{2}$\quad
  Haokun Lin$^{3}$\quad
  Panwen Hu$^{3}$\quad
  Bohang He$^{4}$\quad
  Siyu Ren$^{4}$\\
  Xiaoyu Guo$^{5}$\quad
  Weijia Liufu$^{1}$\quad
  Shuoshuo Xue$^{1}$\quad
  Fengjiao Chen$^{4}$\quad
  Zipeng Ye$^{4}$\\
  Xuezhi Cao$^{4}$\quad
  Zutao Jiang\quad
  Hanhui Li$^{1}$\quad
  Xiaodan Liang$^{1,3,*}$\\[0.8em]
  {\small
  $^{1}$Sun Yat-sen University\quad
  $^{2}$University of Macau\quad
  $^{3}$Mohamed bin Zayed University of Artificial Intelligence}\\
  {\small
  $^{4}$Meituan\quad
  $^{5}$South China University of Technology}\\[0.8em]
  {\normalsize Project page: \href{https://zijunwa.github.io/proworld/}{\texttt{https://zijunwa.github.io/proworld/}}}%
}
\date{}

\begin{document}
\maketitle
{\renewcommand{\thefootnote}{\fnsymbol{footnote}}%
\footnotetext[1]{Corresponding author.}%
\footnotetext[2]{Work done during an internship at Meituan.}}
\begin{abstract}
Evaluating robot manipulation policies through repeated physical trials is costly. We study whether an action-conditioned world model can serve as a closed-loop test environment whose policy success rates track those on real robots. Beyond realistic video, the model must return the camera views and proprioceptive state needed for the policy's next action and preserve failed-action outcomes to avoid overestimating policy capability. We introduce ProWorld, a world model that jointly predicts multi-view video and robot state, then feeds both back to a frozen policy for repeated interaction. ProWorld comprises a shared transformer for joint video--state flow prediction and Shared-Time View Encoding to distinguish cameras while preserving temporal synchronization. To reduce interaction hallucinations that turn missed grasps into successful lifts, we augment training with failed interactions collected in high-fidelity simulation, reducing inflated success rates in the generated environment. ProWorld achieves the highest overall TriWorldBench score among the compared methods. Across three policies and four local manipulation tasks, it attains a Pearson correlation of 0.95 and a Spearman correlation of 0.94 between generated and real success rates. These results support joint visual-proprioceptive feedback and failure-aware training as a basis for world-model-based policy evaluation.
\end{abstract}

\begin{center}
\begin{minipage}{\linewidth}
\makeatletter\def\@captype{figure}\makeatother
\centering
\includegraphics[width=\linewidth]{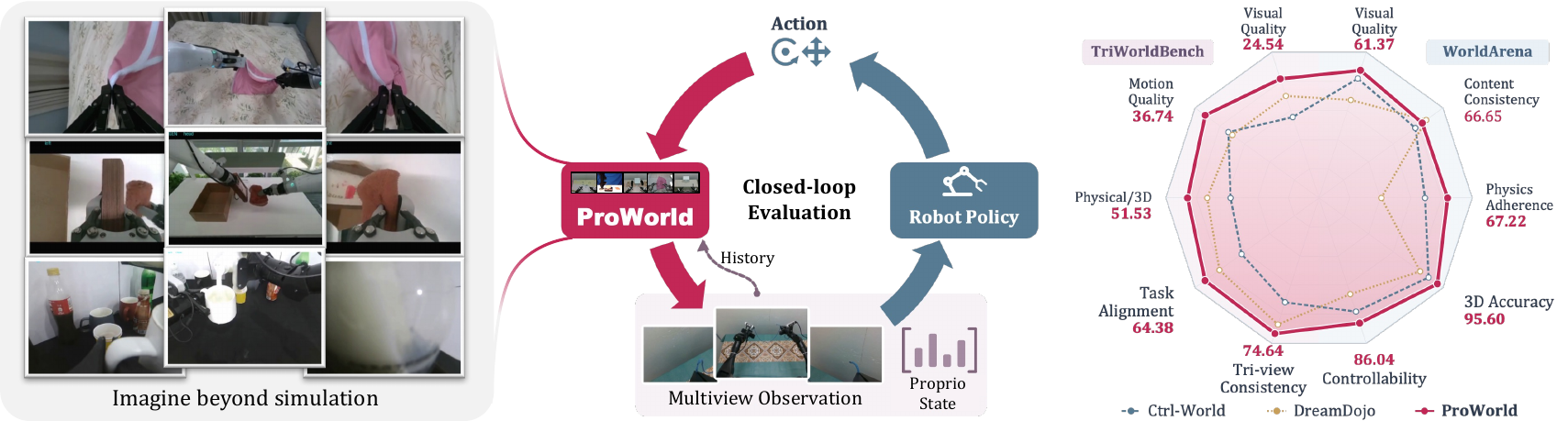}
\caption{\textbf{ProWorld for closed-loop policy evaluation.} Left: ProWorld returns multi-view observations and proprioceptive state to the policy after each action. Right: video-generation performance on TriWorldBench and WorldArena compared with Ctrl-World and DreamDojo.}
\label{fig:teaser}
\end{minipage}
\end{center}

\section{Introduction}
\label{sec:intro}
Evaluating robot policies through physical trials is costly, both in terms of robot time and the need for repeated scene resets. With the development of action-conditioned video foundation models~\citep{unisim,irasim,genieenvisioner,dreamdojo}, recent work has extended these models to policy evaluation~\citep{ctrlworld,gigaworld,roboworld,scalableeval}. These approaches use learned world models as closed-loop evaluation environments: initialized from a real observation, the world model predicts the consequences of actions executed by a fixed policy and generates the observations used for subsequent policy decisions. The goal is to estimate policy success rates through generated closed-loop interactions that faithfully reflect real-robot performance.

However, two requirements that are largely overlooked by existing approaches remain critical for reliable policy evaluation. First, the world model should provide the complete observation space required by the policy, including multiple camera views and proprioceptive states such as joint positions. Second, it must faithfully preserve unsuccessful interactions. For example, if a missed grasp is subsequently rendered as a successful lift, the world model produces an \emph{interaction hallucination}: it predicts a successful outcome that is not supported by the executed action. Such hallucinations can inflate estimated success rates and obscure genuine policy failures. A reliable evaluator therefore requires both complete visual--proprioceptive feedback and faithful modeling of failure outcomes.

We introduce \ours{}, which jointly predicts multi-view videos and robot state from observation history and an action chunk, and then feeds both back to a frozen policy. A shared transformer learns both outputs through joint flow matching. State queries read the visual stream through asymmetric attention, encouraging agreement between predicted state and visible motion. The policy thus receives predicted execution feedback rather than its preceding command as the next state.

To preserve camera identity and synchronization, we develop \emph{Shared-Time View Encoding} (STVE). Independently encoded camera streams receive shared time coordinates and explicit view coordinates. This avoids pixel-tiling boundaries and separates camera identity from elapsed time. Compatible with pretrained video backbones, STVE improves visual prediction over pixel tiling and time concatenation methods.

To reduce interaction hallucinations, we augment demonstrations and robot-policy rollouts with failed interactions collected in high-fidelity simulation. Successful demonstrations provide limited evidence of what happens when actions fail. For example, a missed grasp may leave the object in place, while an unintended push may knock it over. Simulated failures supply this missing supervision. In our local evaluation, this training reduces hallucinated successes and preserves failed interactions.

We assess policy-outcome agreement and video prediction quality. Across three policies and four local tasks, ProWorld achieves a Pearson correlation of 0.95 and a Spearman correlation of 0.94 with real success rates. We also evaluate two policies on six RoboChallenge tasks. On TriWorldBench, ProWorld scores 53.68, exceeding the strongest compared baseline's 51.72.

Our contributions are threefold: \textbf{(1) Joint video--state prediction} that supplies visual observations and learned proprioceptive feedback for closed-loop policy interaction; \textbf{(2) Shared-Time View Encoding}, which preserves camera identity and synchronization while improving visual prediction over pixel tiling and time concatenation; and \textbf{(3) Failure-aware training} that uses simulated failures to reduce interaction hallucinations and success-rate inflation.

\section{Related Work}
\label{sec:related}
\paragraph{Interactive robot world models.}
Generative video models have progressed from visually plausible rollouts to action-conditioned simulators~\citep{unisim,irasim,genieenvisioner,dreamdojo,ctrlworld,gesim2,gigaworld,cosmos3}. Recent systems explicitly use such simulators to assess policies, including scalable video-model evaluation, RoboWorld, and an interactive simulator for long-horizon robot learning~\citep{scalableeval,roboworld,interactivews}. DW0.5 further uses generated successes and failures for policy post-training~\citep{dw05}. Table~\ref{tab:positioning} summarizes these systems. ProWorld combines learned state feedback with calibration-free long-horizon multi-view rollout.

\begin{table}[t]
\centering\scriptsize
\caption{Capabilities of robot world models. ProWorld brings together long-horizon, multi-view generation and proprioceptive feedback without camera calibration or forward kinematics (FK).}
\label{tab:positioning}
\setlength{\tabcolsep}{3pt}
\renewcommand{\arraystretch}{1.12}
\resizebox{\linewidth}{!}{%
\begin{tabular}{@{}lccccccc@{}}
\toprule
\tablehead[l]{Model} & \tablehead{Minute-long\\generation} & \tablehead{Multi-view\\generation} & \tablehead{Calibration-\\free} & \tablehead{FK-free} & \tablehead{State\\output} & \tablehead{Closed-loop\\evaluation} & \tablehead{Multi-embodiment\\adaptation}\\
\midrule
IRASim & $\times$ & $\times$ & $\checkmark$ & $\checkmark$ & $\times$ & $\times$ & $\checkmark$\\
Ctrl-World & $\times$ & $\checkmark$ & $\checkmark$ & $\times$ & $\times$ & $\checkmark$ & $\times$\\
DreamDojo & $\checkmark$ & $\times$ & $\checkmark$ & $\checkmark$ & $\times$ & $\times$ & $\checkmark$\\
GE-Sim 2.0 & $\checkmark$ & $\checkmark$ & $\times$ & $\times$ & $\checkmark$ & $\checkmark$ & $\times$\\
GigaWorld-1 & $\checkmark$ & $\checkmark$ & $\times$ & $\times$ & $\times$ & $\checkmark$ & $\checkmark$\\
DW0.5 & $\times$ & $\checkmark$ & $\checkmark$ & $\checkmark$ & $\times$ & $\times$ & $\checkmark$\\
Xiaomi-Robotics-U0 & $\times$ & $\checkmark$ & $\times$ & $\times$ & $\times$ & $\times$ & $\checkmark$\\
\midrule
\textbf{ProWorld (ours)} & $\checkmark$ & $\checkmark$ & $\checkmark$ & $\checkmark$ & $\checkmark$ & $\checkmark$ & $\checkmark$\\
\bottomrule
\end{tabular}}
\end{table}

\paragraph{Multi-view, multimodal, and world--action models.}
Multiple cameras expose complementary evidence about contact and motion. Multi-View Masked World Models learns visual representations from multiple viewpoints~\citep{mvmwm}, while recent generative systems study joint multi-view prediction and cross-embodiment synthesis~\citep{ctrlworld,adaworldpolicy,xiaomiu0}. Multi-view streams can be organized as pixel canvases~\citep{cosmos3,gigaworld} or concatenated frame sequences~\citep{cosmos25,recammaster}. Geometry-aware rotary embeddings incorporate camera geometry into attention~\citep{geometryrope}; ProWorld instead assigns camera identity a dedicated rotary coordinate alongside shared time, without requiring calibration (Section~\ref{sec:multiview}). World--action models (WAMs) use predictive world representations to produce executable actions: DreamZero jointly models future video and actions, and Motus unifies world modeling, control, and inverse dynamics~\citep{dreamzero,motus}. ProWorld instead supplies the observations through which an external policy is evaluated, including its proprioceptive feedback.

\paragraph{Long-horizon generation and evaluation.}
Diffusion transformers and flow matching provide the modeling substrate for high-capacity video prediction~\citep{dit,flowmatching,wan}; autoregressive diffusion work studies how to continue generated history without excessive drift~\citep{diffusionforcing,selfforcing,causalrcm}. Policy evaluation poses a separate measurement question. WorldArena and TriWorldBench evaluate the quality of action-conditioned video predictions~\citep{worldarena,worldarena2,triworldbench,worldmark}, while real-world and simulation benchmarks evaluate robot policies through task success under specified evaluation protocols~\citep{robochallenge,robotwin2,benchmarks}.

\section{Method}
\label{sec:method}
\subsection{Overview}
\begin{figure}[t]
\centering\includegraphics[width=\linewidth]{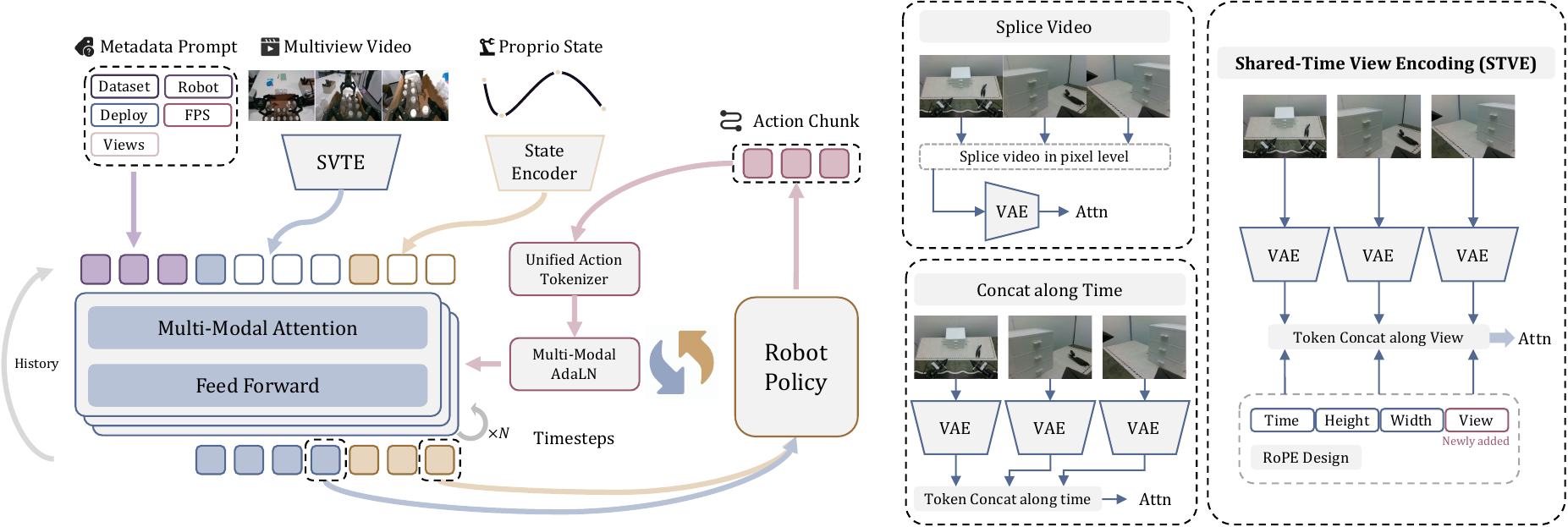}
\caption{\textbf{ProWorld architecture and multi-view organization.} \textbf{Left:} Metadata, independently encoded camera streams, and proprioceptive state enter a shared transformer; action chunks condition visual dynamics through AdaLN, and predicted video and state feed back to the policy. \textbf{Right:} Comparison of three multi-view representations. Pixel tiling combines camera images before VAE encoding, while time concatenation independently encodes each view and concatenates the resulting tokens along the temporal sequence dimension. STVE instead combines independently encoded tokens along the view dimension and uses separate time, height, width, and view rotary coordinates, preserving shared time and per-view spatial structure.}
\label{fig:framework}
\end{figure}

We treat ProWorld as the environment in a repeated observe--act loop, with only the initial observation supplied by the real environment. As shown in Figure~\ref{fig:framework}, at decision time $t$, the policy receives synchronized images $O_t=\{I_t^{(v)}\}_{v=1}^{V}$, proprioceptive state $s_t$, and a task instruction $g$. The policy returns an action chunk $A_t$, and ProWorld predicts the resulting camera observations and robot states over $H$ steps:
\begin{equation}
 A_t\sim\pi(\cdot\mid O_{\leq t},s_{\leq t},g),\qquad
 (\widehat O_{t+1:t+H},\widehat s_{t+1:t+H})\sim
 p_\theta(\cdot\mid O_{\leq t},s_{\leq t},A_t,g).
 \label{eq:loop}
\end{equation}
Here, $O_{\leq t}$ and $s_{\leq t}$ denote the available observation and state histories; actions and observations are aligned by physical time even when their sampling rates differ. The predicted images and state become the policy's next input, and the loop repeats.

\subsection{Joint Video--State Prediction}
\label{sec:joint}
ProWorld jointly predicts visual observations and the resulting robot state using a shared transformer, a joint flow objective, and asymmetric attention. State queries read visual dynamics while both losses train the backbone, providing learned execution feedback for the next policy decision.

\paragraph{Representations and conditioning.}
A video VAE encodes each view into latent tokens. A lightweight projection embeds the normalized proprioceptive state into the transformer width, and a separate output head predicts its flow field. Text and metadata specify the task, embodiment, camera configuration, and sampling rate. Controls are grouped by their associated latent interval and projected into adaptive layer-normalization parameters. State tokens receive their own noise-level embedding; the control modulation is applied to the video tokens.

For end-effector trajectories, we use a velocity-style representation obtained from smoothed poses at the control rate. Let $\widetilde r_k$ contain the continuous pose coordinates and $g_k$ the gripper state. The control feature is
\begin{equation}
 u_k=\left[\frac{\widetilde r_k-\widetilde r_{k-1}}{\Delta t},\;g_k\right],
 \qquad \bar u_k=\mathcal N_u(u_k),\qquad \bar s_k=\mathcal N_s(s_k),
 \label{eq:control}
\end{equation}
where $\mathcal N_u$ and $\mathcal N_s$ are embodiment-specific normalizers. For joint-controlled deployments, the interface uses the policy's joint-command representation and predicts the corresponding measured joint state. 

\begin{figure}[t]
\centering\includegraphics[width=\linewidth]{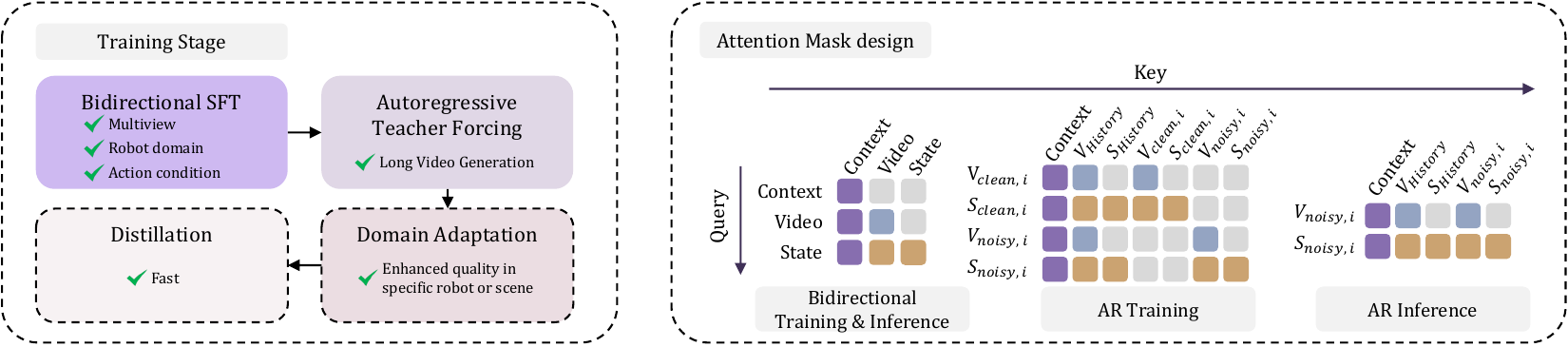}
\caption{\textbf{Training stages and attention visibility.} The training recipe progresses from bidirectional video--state learning to autoregressive teacher forcing, domain adaptation, and distillation. Colored mask entries indicate allowed attention, with queries on rows and keys on columns. Autoregressive training separates clean history from current noisy targets; inference uses generated history.}
\label{fig:stages}
\end{figure}

\paragraph{Joint flow matching.}
Let $x^v$ denote clean video latents and $x^s$ the normalized state sequence. For modality $m\in\{v,s\}$, we interpolate with Gaussian noise $\epsilon^m$ at noise level $\sigma_m$:
\begin{equation}
 x^m_{\sigma_m}=(1-\sigma_m)x^m+\sigma_m\epsilon^m,
 \qquad d^m=\epsilon^m-x^m.
 \label{eq:path}
\end{equation}
The model predicts both flow fields in the same action-conditioned forward pass. We optimize
\begin{equation}
 \mathcal L=\E\!\left[
 w_v(\sigma_v)\|f^v_\theta-d^v\|_{\mathrm{mean}}^2+
 \lambda_s w_s(\sigma_s)\|f^s_\theta-d^s\|_{\mathrm{mean}}^2
 \right],
 \label{eq:loss}
\end{equation}
with clean context excluded from the prediction loss. Here, $\lambda_s$ controls the relative weight of state supervision. We use $\lambda_s=2$ in all experiments with state outputs, except when explicitly varying this weight in Appendix~\ref{app:state-loss-weight}. Sampling integrates from noise to data using the corresponding modality schedules~\citep{flowmatching}. Both losses update the shared transformer.

\paragraph{Attention structure.}
Video and state share a backbone but have different visibility. In bidirectional training, context and video tokens attend to each other; state queries read video and state tokens. Video queries do not read the noisy state stream. This makes state prediction a visual readout while retaining joint supervision of the backbone. For autoregressive training, each noisy chunk reads clean preceding chunks and its own noisy tokens, but not the clean target for that chunk or any future chunk. The state stream also reads preceding states. Figure~\ref{fig:stages} summarizes these masks. During rollout, denoising reads a fixed history cache; the completed chunk refreshes the cache for the next interaction.

\subsection{Shared-Time View Encoding}
\label{sec:multiview}
Figure~\ref{fig:framework} compares three ways to organize multi-view inputs. \textbf{Pixel tiling}, used by Cosmos~3 and GigaWorld-1~\citep{cosmos3,gigaworld}, combines camera images before VAE encoding, reusing the video pipeline but introducing artificial spatial boundaries. \textbf{Time concatenation}, used in Cosmos-Predict2.5's multi-view implementation~\citep{cosmos25} and ReCamMaster's source--target conditioning~\citep{recammaster}, independently encodes each camera stream and concatenates the resulting tokens along the temporal sequence dimension.

\textbf{STVE} instead assigns camera identity its own rotary coordinate (Figure~\ref{fig:framework}, right). Each camera is encoded independently, with tokens indexed by $(t,h,w,v)$: synchronized views share time $t$, $(h,w)$ remain local to each camera, and $v$ identifies the view. Disjoint rotary subspaces encode these coordinates:
\begin{equation}
 \mathcal R(t,h,w,v)=
 \operatorname{diag}\!\left(R_t(t),R_h(h),R_w(w),R_v(v)\right),
 \qquad \widetilde q_i=\mathcal R_i q_i,\quad
 \widetilde k_i=\mathcal R_i k_i.
 \label{eq:rope}
\end{equation}
Joint attention thus distinguishes views without spatial seams or view-dependent time offsets, using the same coordinates for history and generated chunks.

The view coordinate fits within the existing attention-head width: MiniMax-H3 uses its unrotated tail, while Wan reserves a small portion of its spatial rotary channels, preserving the temporal allocation. STVE adds neither tokens nor a separate cross-view attention module. Table~\ref{tab:views} compares the three structures in our model.

\subsection{Failure Data and Closed-Loop Rollout}
\label{sec:failure}
\begin{figure}[!ht]
\centering\includegraphics[width=\linewidth]{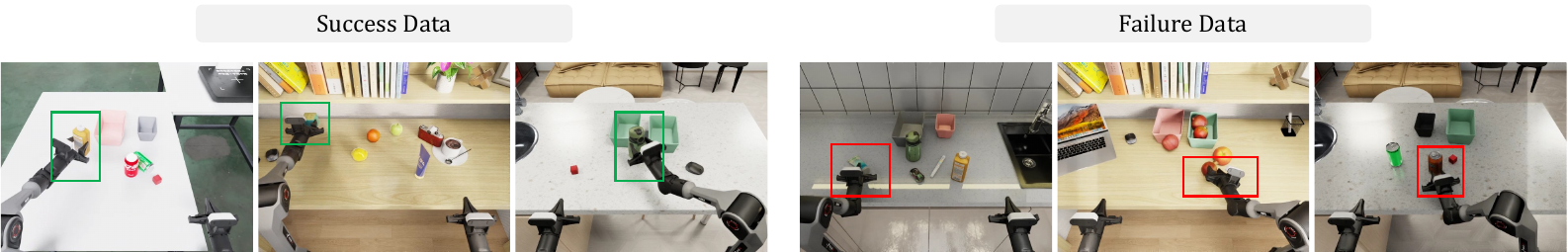}
\caption{\textbf{Success and failure examples in the simulation data.} Successful interactions are shown on the left and failed interactions on the right. Green and red boxes highlight the corresponding robot--object interaction regions.}
\label{fig:failuredataset}
\end{figure}

Successful demonstrations cover intended behavior extensively but offer few examples of policy mistakes. We collect failed trajectories in GenieSim~\citep{geniesim}, a high-fidelity simulator with realistic scenes and physical robot--object interaction, retaining synchronized multi-view observations, commands, and measured states. The observed aftermath of failed actions serves as the prediction target (Figure~\ref{fig:failuredataset}).

Under the same prediction objective, demonstrations supply broad manipulation behavior, robot-policy rollouts capture policy-induced execution, and simulated failures expand unsuccessful interaction coverage. Figure~\ref{fig:agreement} compares generated and physical success rates.

Training follows the four stages in Figure~\ref{fig:stages}. Bidirectional supervised fine-tuning learns action-conditioned video and state prediction. Chunk-wise teacher forcing extends the model to longer rollouts, following an established autoregressive training recipe~\citep{minwm}. Domain adaptation specializes the model to the target robot and observation setup. Following prior work on autoregressive distillation~\citep{causalforcingpp,minwm,helios}, we distill the model for few-step generation. At evaluation time, a frozen policy repeatedly receives generated images and predicted state in its native units, reacting to estimated execution rather than reusing its preceding command. Appendix~\ref{app:benchmarks} provides action/state specifications and temporal alignment, failure-data collection details, and stage-specific training configurations.

\section{Experiments}
\label{sec:experiments}
\begin{center}
\begin{minipage}{\linewidth}
\makeatletter\def\@captype{figure}\makeatother
\centering\includegraphics[width=\linewidth]{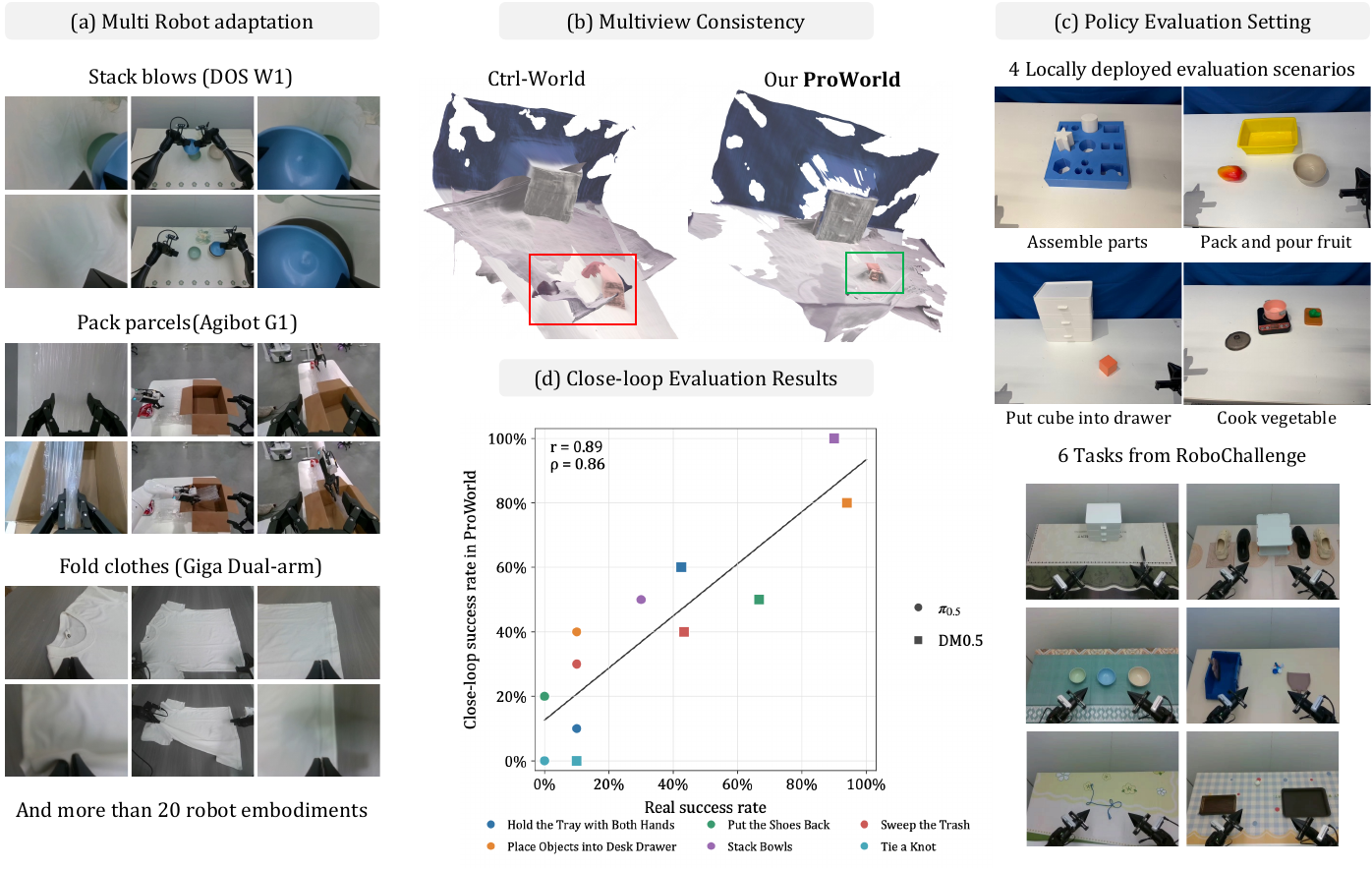}
\caption{\textbf{ProWorld training data and closed-loop evaluation.} (a) Samples from our large-scale multi-embodiment training data. (b) After closed-loop rollout, ProWorld preserves cross-view geometry and reconstructs a consistent scene; Ctrl-World's reconstruction shows a blurred, deformed block. (c) Real-robot evaluation setups and tasks. (d) Generated success rates track physical outcomes on RoboChallenge.}
\label{fig:overviewresults}
\end{minipage}
\end{center}

\subsection{Experimental Setup}
\paragraph{Training data and implementation.}
We pretrain on approximately 6,486 hours of heterogeneous manipulation data, combining large open datasets with a smaller private collection. The mixture includes InternData-A1 (264 h)~\citep{interndata}, RoboCOIN (312 h)~\citep{robocoin}, ABC-130K (2,420 h)~\citep{abc}, InternData-M1 (156 h)~\citep{internm1}, AgiBotWorld-Beta (2,233 h)~\citep{agibot}, RealOmni-Open (199 h)~\citep{realomni}, Galaxea Open-World (119 h)~\citep{galaxea}, and RoboMIND (60 h)~\citep{robomind}, together with 71 hours of GenieSim data containing diverse failure trajectories and 652 hours of private data. We use MiniMax-H3~\citep{h3} and Wan~\citep{wan} as video diffusion backbones, with details on their use and training settings provided in Appendix~\ref{app:training-hyperparameters}.

\paragraph{Evaluation settings.}
The local closed-loop study uses three policies on four manipulation tasks, with Ctrl-World and Cosmos~3 Nano as alternative rollout environments. A second study evaluates two policies on six challenging RoboChallenge tasks~\citep{robochallenge}, covering bimanual handling, placement, sweeping, stacking, and knot tying (Figure~\ref{fig:overviewresults}). For the representation ablations, we initialize models from a pretrained backbone and train them on AgiBotWorld-Alpha~\citep{agibot}. We evaluate action-conditioned video prediction on TriWorldBench~\citep{triworldbench}, comprising 500 three-view trajectories (Table~\ref{tab:triworldbench}), and on WorldArena, comprising 1,000 single-view trajectories.

\begin{table*}[t]
\centering\scriptsize
\caption{Action-conditioned three-view prediction on TriWorldBench. VLM consistency averages the three benchmark VLM scores. Higher is better.}
\label{tab:triworldbench}
\setlength{\tabcolsep}{2.2pt}
\resizebox{\textwidth}{!}{%
\begin{tabular}{@{}lcccccccccc@{}}
\toprule
 & \tablehead{Score} & \multicolumn{2}{c}{\tablehead{Tri-view}} & \tablehead{Task} & \multicolumn{2}{c}{\tablehead{Physical/3D}} & \multicolumn{2}{c}{\tablehead{Motion}} & \tablehead{Temporal} & \tablehead{Visual}\\
\cmidrule(lr){3-4}\cmidrule(lr){6-7}\cmidrule(lr){8-9}
\tablehead[l]{Model} & \tablehead{TWB} & \tablehead{Norm.\\PSNR} & \tablehead{VLM\\consist.} & \tablehead{Instr.\\following} & \tablehead{Interaction\\quality} & \tablehead{Perspective} & \tablehead{State\\alignment} & \tablehead{Trajectory\\accuracy} & \tablehead{Subject\\consist.} & \tablehead{Image\\quality}\\
\midrule
DreamDojo & \underline{51.72} & 59.78 & \underline{79.53} & 49.72 & \textbf{28.28} & 59.39 & \underline{60.76} & \underline{16.61} & 77.37 & \underline{24.45}\\
Motus & 42.35 & \underline{60.13} & 72.82 & \underline{52.28} & 25.00 & 44.20 & 59.31 & 13.77 & 77.37 & 16.08\\
Genie Envisioner & 40.73 & 58.74 & 68.15 & 20.20 & 20.10 & \textbf{87.91} & 59.28 & 0.84 & \textbf{91.40} & 19.36\\
Ctrl-World & 38.98 & 46.39 & 64.20 & 38.90 & 24.19 & 45.15 & 60.47 & 6.09 & 77.05 & 18.37\\
\ours{} & \textbf{53.68} & \textbf{62.55} & \textbf{84.07} & \textbf{62.32} & \underline{28.24} & \underline{74.82} & \textbf{62.01} & \textbf{30.39} & \underline{83.86} & \textbf{26.99}\\
\bottomrule
\end{tabular}}
\end{table*}

\paragraph{Evaluation protocol.}
A frozen policy starts from a real observation, then repeatedly predicts actions from generated images and state. We compare generated and physical success rates using Pearson and Spearman correlations, and measure paired video quality using PSNR, SSIM, and LPIPS~\citep{lpips}. For \emph{video--state consistency error} (VSC), we train DreamGen's inverse-dynamics model (IDM)~\citep{dreamgen} on AgiBotWorld-Alpha and subtract its reference-video position error (approximately 9.1 mm) from the discrepancy between video-inferred and predicted trajectories. VSC is reported in millimeters, alongside motion-direction cosine similarity (Dir.) and gripper accuracy (Grip.). Evaluation details appear in Appendices~\ref{app:closed-loop-configuration} and~\ref{app:trajectory-quality}.

\subsection{Closed-Loop Policy Evaluation}
\begin{figure}[t]
\centering\includegraphics[width=\linewidth]{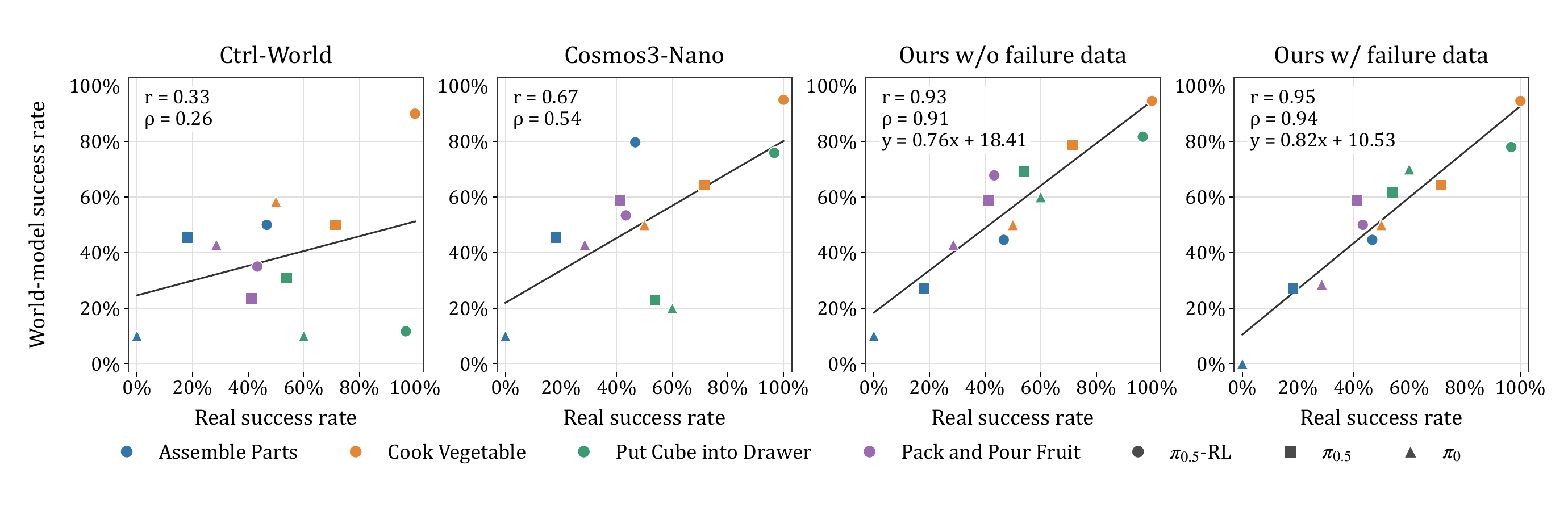}
\caption{\textbf{Policy success in the real and generated worlds.} Colors identify four tasks; shapes identify three policies. Lines are linear fits in percentage units; panels report Pearson $r$ and Spearman $\rho$. Failure-data training improves outcome agreement.}
\label{fig:agreement}
\end{figure}

ProWorld's generated success rates closely track real outcomes across 12 local policy--task pairs (Figure~\ref{fig:agreement}), with Pearson $r=0.95$ and Spearman $\rho=0.94$. Both correlations exceed those of Ctrl-World ($0.33/0.26$) and Cosmos~3 Nano ($0.67/0.54$), indicating stronger agreement in success rates and outcome ordering.

Failure data improves Pearson/Spearman correlation from $0.93/0.91$ to $0.95/0.94$. The fitted relation changes from $y=0.76x+18.41$ to $y=0.82x+10.53$, reducing overestimation at low real success rates. This supports transferring failure supervision across embodiments.

RoboChallenge tasks often require over half a minute of stable closed-loop execution, with coherent observations throughout object manipulation. ProWorld supports these sustained interactions and preserves success-rate variation across policies and tasks (Figure~\ref{fig:overviewresults}d), extending evaluation beyond short demonstrations of a single successful motion.

\begin{figure}[t]
\centering\includegraphics[width=0.95\linewidth]{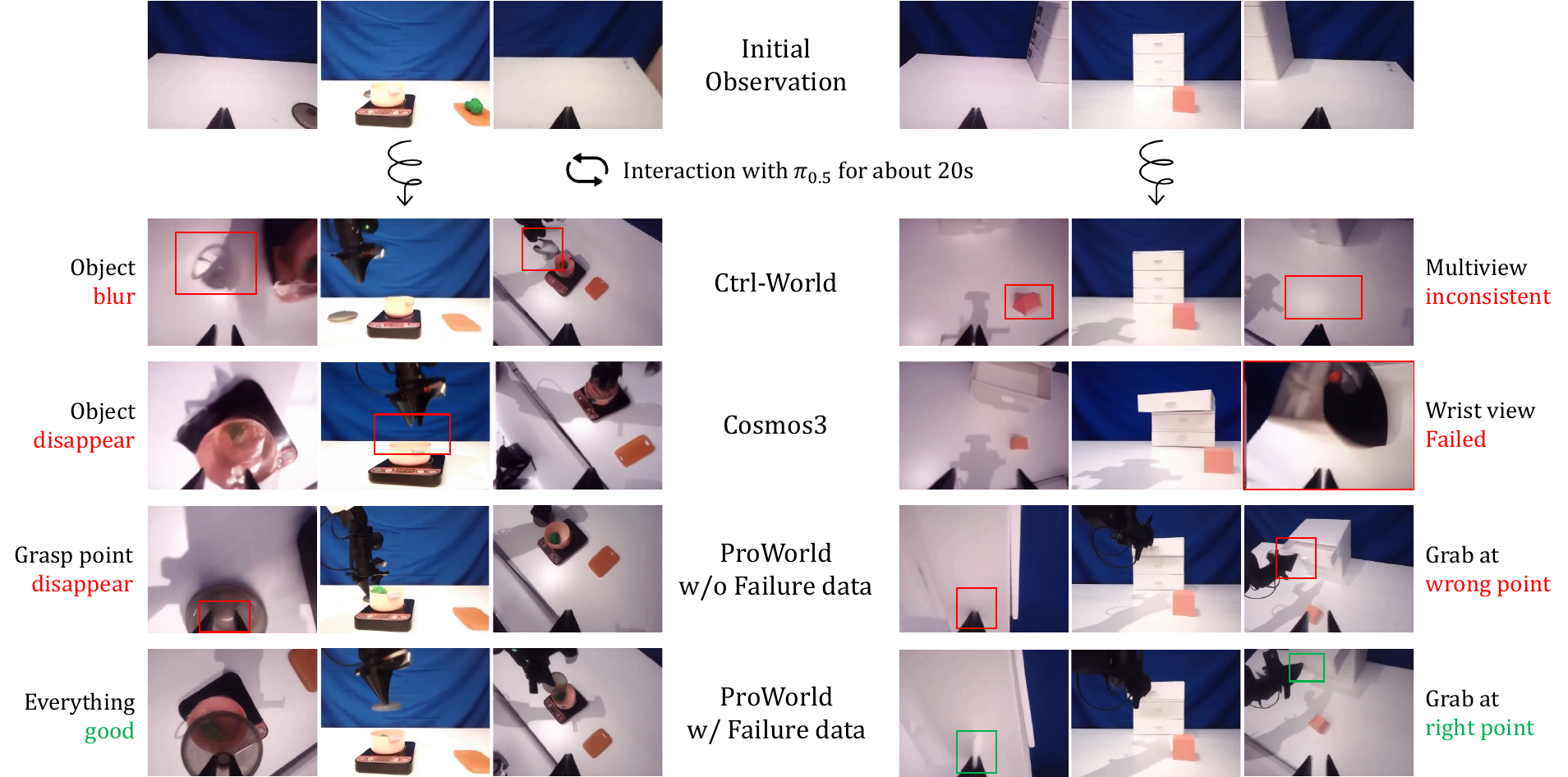}
\caption{\textbf{Closed-loop visual comparison after roughly 20 seconds.} Ctrl-World, Cosmos~3 Nano, and ProWorld with and without failure data start from matched observations. Boxes highlight disappearing objects, misplaced grasps, and improved contact consistency with the full model.}
\label{fig:viscompare}
\end{figure}

\subsection{Video Prediction and Multi-View Quality}
On TriWorldBench (Table~\ref{tab:triworldbench}), ProWorld achieves the highest overall score, with gains in cross-view agreement, task following, and motion. Head and wrist observations provide complementary constraints on robot motion and scene interaction.

On the single-view WorldArena benchmark, ProWorld performs especially well on trajectory accuracy, photometric consistency, and semantic alignment (Table~\ref{tab:worldarena}). Together, these results indicate that motion, persistent scene appearance, and task content survive the generated transition. Appendix~\ref{app:benchmark-results} reports detailed results for both benchmarks and explains the exclusion of frame-rate-sensitive WorldArena motion-quality metrics and EWMScore.

\begin{table*}[t]
\centering\scriptsize
\caption{Single-view prediction on WorldArena Track~1. Higher is better; bold and underline mark the best and second-best results.}
\label{tab:worldarena}
\setlength{\tabcolsep}{2.2pt}
\resizebox{\textwidth}{!}{%
\begin{tabular}{@{}lcccccccccc@{}}
\toprule
 & \multicolumn{2}{c}{\tablehead{Visual}} & \multicolumn{2}{c}{\tablehead{Consistency}} & \multicolumn{2}{c}{\tablehead{Physics}} & \multicolumn{2}{c}{\tablehead{3D}} & \multicolumn{2}{c}{\tablehead{Control}}\\
\cmidrule(lr){2-3}\cmidrule(lr){4-5}\cmidrule(lr){6-7}\cmidrule(lr){8-9}\cmidrule(l){10-11}
\tablehead[l]{Model} & \tablehead{Image\\quality} & \tablehead{JEPA\\similarity} & \tablehead{Subject\\consist.} & \tablehead{Photo.\\consist.} & \tablehead{Interaction\\quality} & \tablehead{Trajectory\\accuracy} & \tablehead{Depth\\accuracy} & \tablehead{Perspective} & \tablehead{Instr.\\following} & \tablehead{Semantic\\alignment}\\
\midrule
Ctrl-World & 42.44 & 92.77 & \textbf{83.56} & 12.88 & 62.62 & 48.20 & 93.25 & 83.66 & 67.68 & 88.68\\
GigaWorld-1 & 51.18 & 96.77 & 80.97 & \underline{28.11} & 75.10 & 54.27 & \textbf{98.44} & 95.60 & 82.14 & 89.42\\
ABot-PhysWorld & \textbf{61.03} & 90.36 & 80.56 & 20.52 & \underline{81.96} & 31.50 & 71.99 & \textbf{98.94} & \underline{92.10} & \underline{89.58}\\
GE-Sim 2.0 & 46.73 & \underline{97.16} & \underline{82.50} & 7.05 & 73.60 & \underline{58.63} & 96.49 & 93.64 & 82.70 & 89.05\\
Xiaomi-Robotics-U0 & \underline{53.94} & 90.60 & 79.05 & 2.13 & \textbf{87.30} & 41.89 & 85.25 & \underline{98.84} & \textbf{93.86} & 89.35\\
\ours{} & 45.82 & \textbf{98.39} & 78.91 & \textbf{35.34} & 72.25 & \textbf{62.19} & \underline{98.36} & 92.85 & 81.66 & \textbf{90.41}\\
\bottomrule
\end{tabular}}
\end{table*}

\paragraph{What changes during interaction?}
Figure~\ref{fig:viscompare} compares two tasks from matched initial scenes. Cooking requires retaining the pot and its graspable parts; drawer manipulation requires consistent gripper contact across wrist and workspace views. Baselines exhibit object disappearance, blurred structure, and inconsistent views. ProWorld preserves these details, giving the policy stable visual context for its next decision.

\begin{figure}[t]
\centering\includegraphics[width=\linewidth]{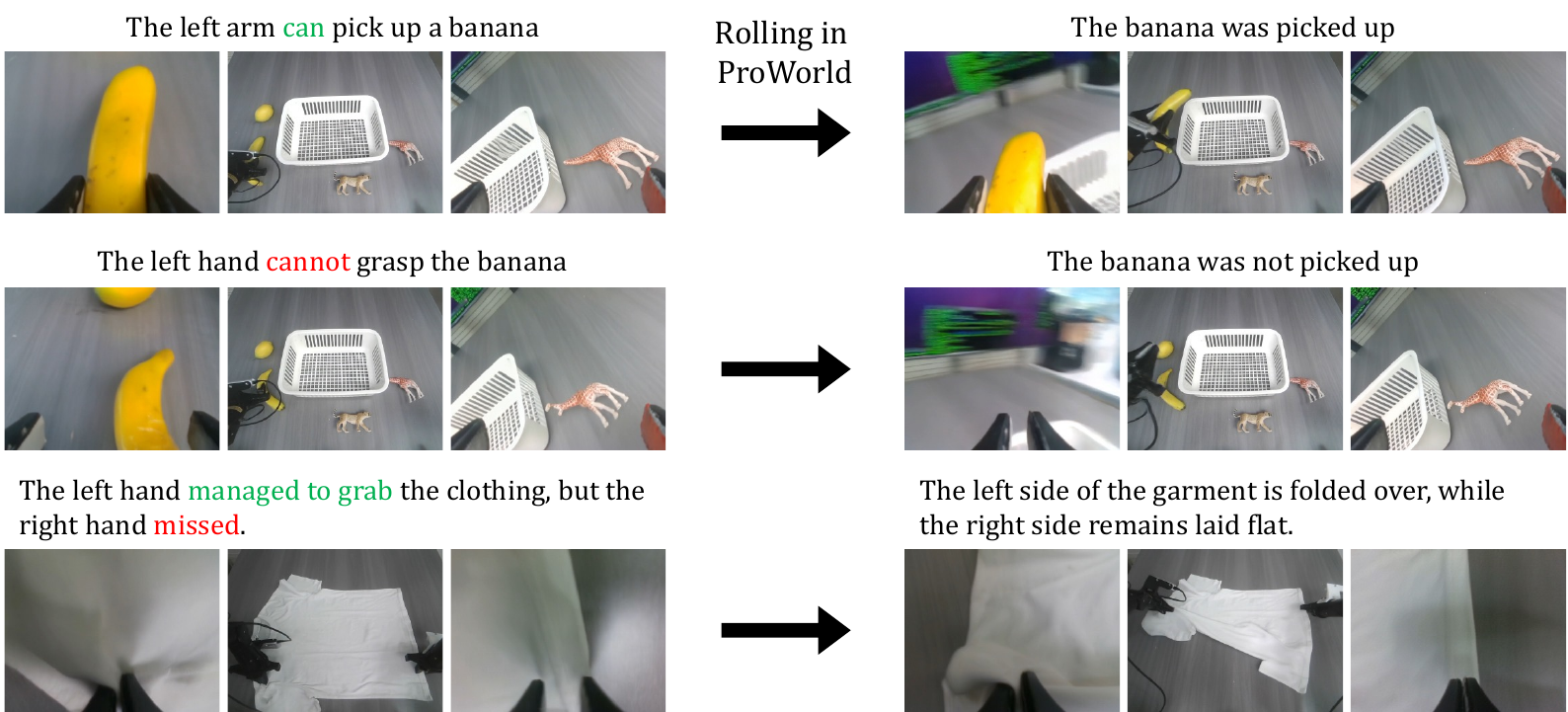}
\caption{\textbf{Generated three-view successes and failures.} Rows: successful and missed banana pickups, then an incomplete cloth fold. Right: \ours{} continuations after the actions at left.}
\label{fig:successfailure}
\end{figure}

Without failure data, errors concentrate around contact: handles, grasp points, and gripper--object positions. Failure examples teach the model how scenes evolve when contact is absent or incomplete. In Figure~\ref{fig:successfailure}, an empty lift leaves the banana behind, and a one-sided grasp produces a one-sided fold. These distinctions lead to different closed-loop task outcomes.

\subsection{Ablations}
Further ablations of action conditioning, visual context, noise schedules, and training resolution are provided in Appendix~\ref{app:h3-ablations}.
\paragraph{Action representation and joint state learning.}
In Table~\ref{tab:action}, velocity-style control improves visual prediction over absolute-state conditioning by directly relating the commanded change to visible motion. Adding state supervision improves the shared visual representation and supplies proprioceptive feedback. Motion-based conditioning with asymmetric attention gives the strongest combination: state queries read the evolving video, while state loss trains the backbone to retain robot-motion information.

Compared with full attention between noisy streams, asymmetric attention lowers VSC and improves motion direction with little change in gripper accuracy, thereby strengthening agreement between predicted continuous motion and the generated video.

\begin{table}[t]
\centering\footnotesize
\caption{Control, prediction target, and attention ablation on AgiBotWorld-Alpha.}
\label{tab:action}
\setlength{\tabcolsep}{3pt}
\begin{tabular*}{\linewidth}{@{\extracolsep{\fill}}llcccccc@{}}
\toprule
 &  & \multicolumn{3}{c}{\tablehead{Video quality}} & \multicolumn{3}{c}{\tablehead{State quality}}\\
\cmidrule(lr){3-5}\cmidrule(l){6-8}
\tablehead[l]{Control} & \tablehead[l]{Targets} & \tablehead{PSNR$\uparrow$} & \tablehead{SSIM$\uparrow$} & \tablehead{LPIPS$\downarrow$} & \tablehead{VSC$\downarrow$} & \tablehead{Dir.$\uparrow$} & \tablehead{Grip.$\uparrow$}\\
\midrule
Absolute state & Video & 21.21 & 0.798 & 0.1342 & \NA & \NA & \NA\\
Velocity & Video & 22.18 & 0.819 & 0.1167 & \NA & \NA & \NA\\
Absolute state & Video + state & 21.45 & 0.806 & 0.1130 & 20.5 & 0.380 & \textbf{0.976}\\
Velocity & Video + state (full attn.) & 22.47 & 0.821 & \textbf{0.1053} & 20.3 & 0.408 & 0.974\\
Velocity & Video + state (ours) & \textbf{22.60} & \textbf{0.829} & 0.1069 & \textbf{14.2} & \textbf{0.464} & 0.974\\
\bottomrule
\end{tabular*}
\end{table}

\paragraph{Multi-view encoding.}
Table~\ref{tab:views} compares the three structures in Figure~\ref{fig:framework}. Time concatenation improves all video metrics over pixel tiling. Replacing its view-dependent temporal offsets with STVE further raises PSNR from 21.56 to 22.60 and SSIM from 0.809 to 0.829, and lowers LPIPS from 0.1266 to 0.1069. State results are mixed: STVE improves directional agreement and matches gripper accuracy, but time concatenation has lower VSC. Thus, STVE benefits visual prediction rather than uniformly improving all outputs.

\begin{table}[!ht]
\centering\footnotesize
\caption{Multi-view encoding ablation. State metrics follow Table~\ref{tab:action}.}
\label{tab:views}
\setlength{\tabcolsep}{3pt}
\begin{tabular*}{\linewidth}{@{\extracolsep{\fill}}lcccccc@{}}
\toprule
 & \multicolumn{3}{c}{\tablehead{Video quality}} & \multicolumn{3}{c}{\tablehead{State quality}}\\
\cmidrule(lr){2-4}\cmidrule(l){5-7}
\tablehead[l]{Encoding} & \tablehead{PSNR$\uparrow$} & \tablehead{SSIM$\uparrow$} & \tablehead{LPIPS$\downarrow$} & \tablehead{VSC$\downarrow$} & \tablehead{Dir.$\uparrow$} & \tablehead{Grip.$\uparrow$}\\
\midrule
Pixel tiling & 17.95 & 0.619 & 0.2843 & 23.2 & 0.318 & 0.969\\
Time concatenation & 21.56 & 0.809 & 0.1266 & \textbf{13.3} & 0.416 & \textbf{0.974}\\
STVE (ours) & \textbf{22.60} & \textbf{0.829} & \textbf{0.1069} & 14.2 & \textbf{0.464} & \textbf{0.974}\\
\bottomrule
\end{tabular*}
\end{table}

\section{Conclusion and Limitations}
ProWorld evaluates robot policies through closed-loop interaction, combining joint visual--proprioceptive feedback, STVE for camera identity and synchronization, and simulated failures to reduce interaction hallucinations and success-rate inflation. Strong correlation with real policy success and competitive video prediction support complete observation feedback and faithful failure outcomes as foundations for policy evaluation.

Object deformation and drift over exceptionally long rollouts remain limitations. Improving geometry persistence and rollout stability could reduce accumulated evaluation errors.

\bibliography{references}
\bibliographystyle{iclr2027_conference}

\clearpage
\appendix
\begin{center}
{\Large\bfseries Appendix}
\end{center}
\section{Training and Interface Details}
\label{app:benchmarks}

\subsection{Training Details}
\label{app:training-hyperparameters}

\paragraph{Backbones and initial fine-tuning.}
We instantiate \ours{} with two pretrained video backbones: Wan2.1-14B and MiniMax-H3. We perform full-parameter fine-tuning on Wan2.1-14B and low-rank adaptation (LoRA) on MiniMax-H3, using learning rates of $2\times10^{-5}$ and $10^{-4}$, respectively. Both models are initially fine-tuned on our proposed dataset for 36,000 steps with a batch size of 64.

\paragraph{Wan-based training and experiments.}
Following this initial fine-tuning, we continue training the Wan-based model using the standard Helios Stage~1 procedure with an autoregressive attention mask. The ablation studies in the main text and the RoboChallenge experiments use this Wan-based model.

\paragraph{MiniMax-H3-based training and experiments.}
For MiniMax-H3, we follow the standard minWM pipeline for long-video generation training and distillation after the initial fine-tuning. The additional ablation studies in Appendix~\ref{app:h3-ablations} use the MiniMax-H3-based model.

\subsection{Action/State Definitions and Temporal Alignment}
\label{app:action-state-alignment}

\paragraph{Definitions and layouts.}
The control feature $u_k$ conditions prediction, whereas $s_k$ is the absolute proprioceptive target. End-effector data use 32-dimensional vectors: each arm has Cartesian position (3), a 6D rotation representation (6), and gripper opening (1), occupying entries 0--9 and 10--19; unused entries are zero. Continuous pose coordinates are median-filtered and Savitzky--Golay smoothed before differencing at the control rate. Translation features have units of length per second; rotation features are 6D-coordinate rates, not angular velocities. Grippers retain raw absolute openings. State targets remain the original absolute poses and openings. These trajectory-derived controls are motion proxies. For joint data with command logs, absolute commanded and measured positions remain separate. The RoboTwin adapter, for example, places each arm's six joints and gripper in entries 0--6 and 16--22, with zeros elsewhere.

\paragraph{Normalization.}
$\mathcal N_u$ and $\mathcal N_s$ follow the dataset/embodiment schema. Scalar coordinates use 1st/99th-percentile affine scaling; absolute 6D rotations retain their bounded representation. Velocity features use separate statistics, with continuous coordinates clipped to $[-5,5]$ and grippers to $[-1,1]$ after normalization. Joint data use joint-layout statistics. State predictions are mapped back to native units before policy feedback; clipping is not invertible.

\paragraph{Temporal alignment.}
Synchronized views share source-frame indices. For clip start $b$, source rate $f_s$, video rate $f_v$, and control rate $f_u$, the binary loader samples
\begin{equation}
 i_j^{v}=b+\operatorname{round}(j f_s/f_v),\qquad
 i_k^{u}=b+\operatorname{round}(k f_s/f_u).
\end{equation}
The initial image uses $j=0$; target controls start at $k=1$. Each state target is read at its control's source index. At equal video/control rates, $(u_k,s_k)$ aligns with target image $O_k$. For $k>1$, the derived control describes the interval ending at that sample. The first sampled velocity of each continuous coordinate is zero; the gripper retains its absolute value. Across rates, alignment follows physical time rather than equal array positions. History uses the same sampling grid, with control indices before the episode start clamped to its initial sample.

\subsection{Metadata Prompt Format}
\label{app:metadata-prompt}
\paragraph{Fields and serialization.}
The metadata prompt describes the data source, embodiment, and observation setup in a compact JSON object. Its fields have a fixed order: \texttt{dataset} identifies the dataset group or individual data subset; \texttt{robot} names the embodiment; \texttt{fps} gives the control sampling rate in Hz; \texttt{deploy} specifies the domain (e.g., \texttt{real} or \texttt{simulation}); and \texttt{views} lists camera names in the same order as the input video streams. Thus, \texttt{fps} corresponds to $f_u$ in Appendix~\ref{app:action-state-alignment}, and the camera list is reordered whenever the views are shuffled.

\paragraph{Example and optional task text.}
For AgiBotWorld-Alpha at 10\,Hz, group-level metadata take the following form (line breaks are added for readability):
{\small
\begin{verbatim}
{"dataset":"agibot/agibot_alpha","robot":"agibot_alpha",
 "fps":10,"deploy":"real",
 "views":["head_color","hand_left_color","hand_right_color"]}
\end{verbatim}
}
The loader serializes this object without extra whitespace. Group-level naming permits samples with the same metadata to share a prompt; subset-level naming retains the individual dataset identifier while removing a trailing train/evaluation split name. If task conditioning is enabled and a nonempty instruction is available, a final \texttt{task} string is appended to the object. Otherwise, that field is omitted. Numeric actions and proprioceptive targets are supplied through the separate interface in Appendix~\ref{app:action-state-alignment}, rather than inserted into this JSON object.

\subsection{Failure Data Collection}
\label{app:failure-data-collection}

\paragraph{Perturbed physical replay.}
We generate missed-grasp examples from successful GenieSim trajectories. Each replay restores the recorded scene and initial robot--object poses, then tracks the recorded joint trajectory with a Cartesian offset applied to the active arm. A damped least-squares Jacobian update maps the offset to joint corrections, ramped in before gripper closure and maintained during lifting. The target object remains a dynamic rigid body, allowing contact and gravity to determine the outcome; non-target objects remain kinematic.

\paragraph{Failure selection and recording.}
We label a perturbed replay as a missed grasp only when the target's maximum vertical displacement from its first replay frame is below a threshold adjusted during collection and bounded by 5\,cm. Replays reaching this threshold are excluded from the failure set. Each replay saves synchronized camera videos, the perturbed joint reference trajectory, and per-frame target and end-effector poses. A manifest links each variant to its successful source and records the requested offset, achieved offset at closure, and lift-based label.

\subsection{Closed-Loop Evaluation Configuration}
\label{app:closed-loop-configuration}

\paragraph{Adaptation and policy prerequisites.}
Before closed-loop evaluation, we fine-tune \ours{} for the target embodiment and tasks using trajectories collected on the corresponding real-robot platform. GenieSim failure trajectories constitute 10\% of the fine-tuning mixture. Evaluation additionally requires locally deployable robot-policy weights and real-robot success rates for the same policies and tasks.

\paragraph{Closed-loop protocol.}
Evaluation recordings provide only the initial frames for world-model rollouts; subsequent recorded video and actions are never supplied to either \ours{} or the policy. After initialization, the policy acts on observations generated by \ours{}, which predicts the next multi-view observations and proprioceptive state for continued policy feedback. Thus, all subsequent actions are chosen online by the policy. Below, one interaction round denotes one policy query followed by the corresponding world-model transition.

\paragraph{RoboChallenge.}
We use the Table30v2 dataset for adaptation and evaluate $\pi_{0.5}$ and DM0.5, selected for their publicly released fine-tuned policy weights and nontrivial success rates reported on the official test platform. We use these official rates as the real-robot reference and collect 120 evaluation trajectories from the RoboChallenge website across the two policies and six tasks. These recordings supply initial frames under the protocol above. Most tasks require more than 80 interaction rounds to complete; success in the world-model rollouts is judged manually.

\paragraph{Cobot.}
For our local Cobot evaluation, we collect 200 real-robot trajectories across three policies and four tasks. These recordings serve only to measure real-robot policy success rates and provide initial frames for world-model evaluation. Cobot tasks require approximately 25 policy--world-model interaction rounds. A locally deployed vision--language model (VLM) scores the generated rollouts to determine task success.

\section{Additional Ablation Studies}
\label{app:h3-ablations}

We examine action conditioning, the state-loss weight, visual context, noise schedules, and training resolution. The action-conditioning, visual-context, noise-schedule, and training-resolution ablations use MiniMax-H3 on AgiBotWorld-Alpha, with 108 evaluation clips per setting. Slight differences in the starting configurations make absolute scores across ablations not directly comparable.

\subsection{Trajectory-Quality Evaluation Protocol}
\label{app:trajectory-quality}
\paragraph{Inputs and alignment.}
The AgiBotWorld-Alpha ablations assess video--state agreement and state prediction separately. For video--state agreement, a fixed inverse dynamics model (IDM) reads only the head-camera video at 10 FPS, resized to $224\times224$. We remove the initial conditioning frame to align video frames with saved state predictions. The IDM takes pairs of frames 16 steps apart and predicts 16 motion steps; overlapping predictions are averaged and denormalized. Cumulative motion is anchored at the first state of the trajectory being compared, while gripper outputs remain absolute.

\paragraph{Video--state consistency (VSC).}
Let $E(V,S)$ denote the mean Euclidean distance, in millimeters, between the IDM-reconstructed end-effector positions from video $V$ and the positions in state sequence $S$. We average over time and both arms within each clip, then equally over clips. With generated video $\hat V$, predicted states $\hat S$, and recorded references $(V,S)$, we report
\begin{equation}
\mathrm{VSC}=E(\hat V,\hat S)-E(V,S).
\end{equation}
Thus, VSC measures excess video--state disagreement relative to the IDM's reference-video error; lower is better. This reference error is an empirical calibration, not a guaranteed lower bound. VSC covers the head view only and does not measure task success.

\paragraph{State prediction (Dir.\ and Grip.).}
Dir.\ is the motion-direction cosine similarity between predicted and ground-truth end-effector trajectories. Grip.\ is the accuracy of predicted gripper open/closed states against ground truth. Both are higher-is-better measures computed from predicted versus reference states, independently of the IDM. State metrics are omitted for models without state supervision.

\subsection{Action Conditioning and State Supervision}
Direct AdaLN provides a strong interface between actions and visual dynamics (Table~\ref{tab:h3-core}). Replacing it with velocity tokens reduces PSNR by 3.48 dB and increases LPIPS from 0.1698 to 0.2786. Action guidance recovers part of this gap, but direct modulation remains substantially stronger. State supervision also benefits visual prediction: removing it costs 1.36 dB and degrades both structural and perceptual quality.

\begin{table}[!htbp]
\centering\footnotesize
\caption{\textbf{Action conditioning and state supervision.} Direct AdaLN improves visual prediction over token-based conditioning.}
\label{tab:h3-core}
\setlength{\tabcolsep}{4pt}
\begin{tabular}{@{}lccc@{}}
\toprule
\tablehead[l]{Variant} & \tablehead{PSNR$\uparrow$} & \tablehead{SSIM$\uparrow$} & \tablehead{LPIPS$\downarrow$}\\
\midrule
Direct AdaLN & \textbf{19.03} & \textbf{0.747} & \textbf{0.1698}\\
\quad w/o state supervision & 17.67 & 0.720 & 0.1996\\
Velocity tokens & 15.55 & 0.672 & 0.2786\\
Velocity tokens with action CFG & 16.23 & 0.699 & 0.2422\\
\bottomrule
\end{tabular}
\end{table}

\subsection{State-Loss Weight}
\label{app:state-loss-weight}
\begin{figure}[!htbp]
\centering\includegraphics[width=\linewidth]{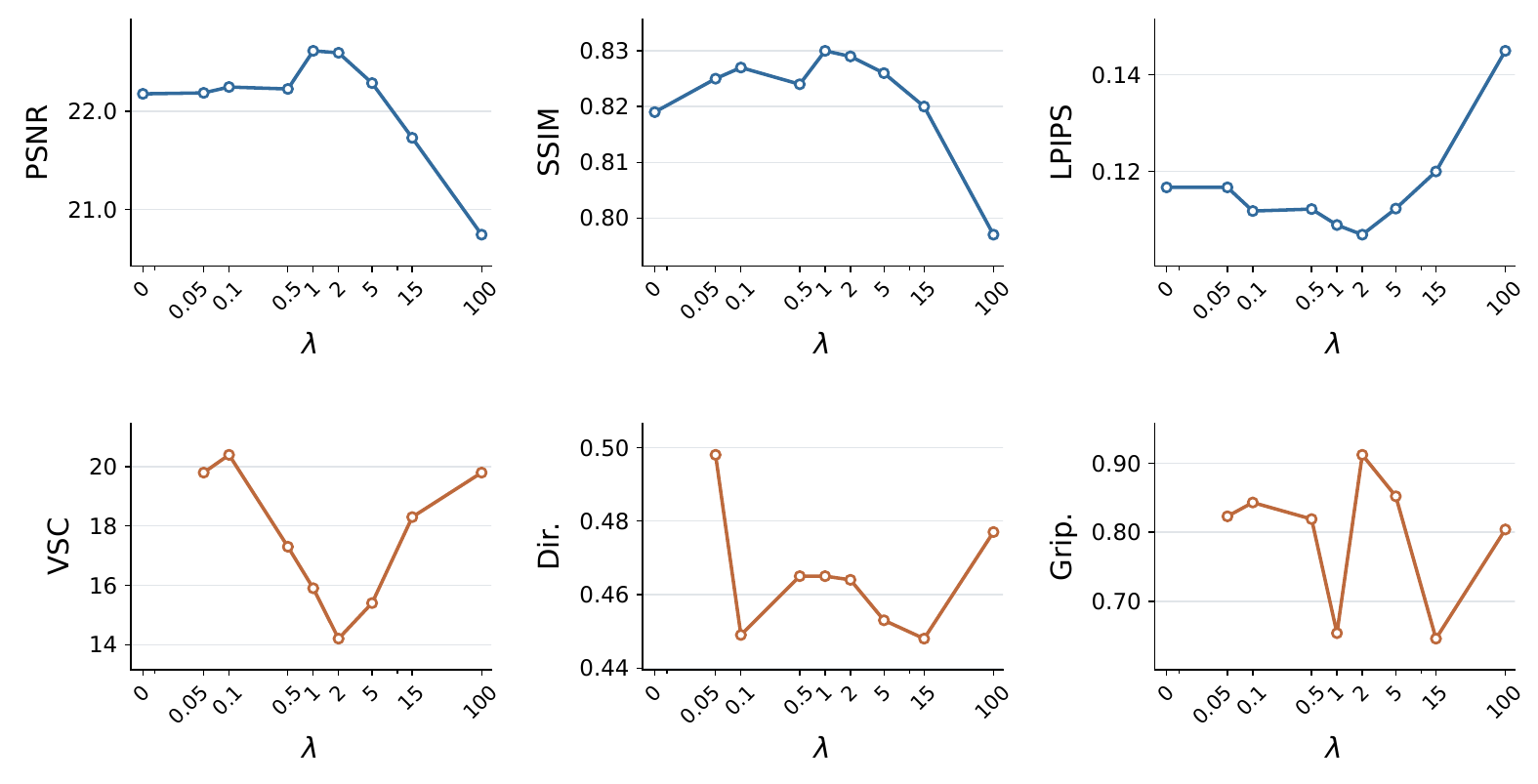}
\caption{\textbf{Ablation of the state-loss weight.} The horizontal-axis label $\lambda$ denotes $\lambda_s$ in the main text. Higher PSNR (dB), SSIM, motion-direction cosine similarity (Dir.), and gripper accuracy (Grip.) are better; lower LPIPS and video--state consistency error (VSC, mm) are better. State-related metrics are shown only for nonzero weights. The default setting is $\lambda_s=2$.}
\label{fig:state-lambda-ablation}
\end{figure}

The state-loss weight controls how strongly proprioceptive supervision contributes to joint video--state learning. Figure~\ref{fig:state-lambda-ablation} varies $\lambda_s$, the coefficient of the state term in the main text's joint flow-matching objective, over $\{0,0.05,0.1,0.5,1,2,5,15,100\}$. The figure uses $\lambda$ as shorthand for $\lambda_s$; $\lambda_s=0$ removes state supervision, and state-related metrics are omitted for this setting.

A moderate state-loss weight provides a favorable balance between visual quality and video--state agreement. Among the tested values, $\lambda_s=2$ yields the lowest LPIPS and VSC and the highest gripper accuracy, while PSNR and SSIM are slightly higher at $\lambda_s=1$. Motion-direction similarity does not peak at $\lambda_s=2$, so this choice does not optimize every metric individually. Larger weights, particularly 15 and 100, degrade visual quality and increase VSC relative to 2. We therefore adopt $\lambda_s=2$ for all experiments with state outputs, except for this explicit weight sweep.

\subsection{Visual Context and Noise Scheduling}
Initial scene descriptions and noise scheduling improve different aspects of prediction (Table~\ref{tab:h3-shift}). Descriptions derived from synchronized initial views reduce video--state consistency error from 9.71 to 7.93 mm, with little change in visual quality. Increasing the shared video/state flow shift to 12 gives the strongest visual scores, improving PSNR by 0.43 dB and reducing LPIPS to 0.1472. The lower shift of 5 preserves better motion direction and video--state agreement, revealing a trade-off between visual fidelity and motion accuracy.

\begin{table}[!htbp]
\centering\footnotesize
\caption{\textbf{Visual context and training flow shift.} Video and state use the same shift. VSC is measured in mm.}
\label{tab:h3-shift}
\setlength{\tabcolsep}{5pt}
\begin{tabular*}{\textwidth}{@{\extracolsep{\fill}}lcccccc@{}}
\toprule
\tablehead[l]{Variant} & \tablehead{PSNR$\uparrow$} & \tablehead{SSIM$\uparrow$} & \tablehead{LPIPS$\downarrow$} & \tablehead{VSC$\downarrow$} & \tablehead{Dir.$\uparrow$} & \tablehead{Grip.$\uparrow$}\\
\midrule
Default & 19.88 & 0.757 & 0.1536 & 9.71 & \textbf{0.443} & \textbf{0.996}\\
Initial multi-view descriptions & 19.73 & 0.756 & 0.1553 & \textbf{7.93} & 0.437 & \textbf{0.996}\\
Shared flow shift 5 & 20.02 & 0.758 & 0.1523 & 8.24 & 0.425 & \textbf{0.996}\\
Shared flow shift 12 & \textbf{20.31} & \textbf{0.762} & \textbf{0.1472} & 8.64 & 0.407 & 0.995\\
\bottomrule
\end{tabular*}
\end{table}

\subsection{Coordinating Video and State Denoising}
Separate noise schedules let video and state prediction use different flow shifts while sharing the same denoising progress. Table~\ref{tab:h3-clock} compares a shared shift of 12 with separate video/state shifts of 12/5, using either aligned or independently sampled progress. Aligned progress gives the lowest VSC and highest motion-direction similarity, while maintaining the visual quality of the shared schedule. Independent sampling raises VSC from 7.83 to 8.91 mm. Coordinating denoising progress therefore matters even when the two modalities use different noise levels.

\begin{table}[!htbp]
\centering\footnotesize
\caption{\textbf{Video and state noise schedules.} Shift pairs denote video/state shifts; aligned progress shares a sampled denoising index across the two schedules.}
\label{tab:h3-clock}
\setlength{\tabcolsep}{5pt}
\begin{tabular*}{\textwidth}{@{\extracolsep{\fill}}lcccccc@{}}
\toprule
\tablehead[l]{Variant} & \tablehead{PSNR$\uparrow$} & \tablehead{SSIM$\uparrow$} & \tablehead{LPIPS$\downarrow$} & \tablehead{VSC$\downarrow$} & \tablehead{Dir.$\uparrow$} & \tablehead{Grip.$\uparrow$}\\
\midrule
12/12, shared schedule & \textbf{20.54} & \textbf{0.765} & \textbf{0.1438} & 8.38 & 0.422 & \textbf{0.996}\\
12/5, aligned progress & 20.51 & 0.764 & 0.1445 & \textbf{7.83} & \textbf{0.440} & 0.995\\
12/5, independent progress & 20.38 & 0.763 & 0.1458 & 8.91 & 0.430 & 0.995\\
\bottomrule
\end{tabular*}
\end{table}

\subsection{Mixed-Resolution Training}
Mixed-resolution training substantially improves 480p prediction (Table~\ref{tab:h3-resolution}). Compared with training exclusively at 480p, it raises PSNR by 2.11 dB, lowers VSC by 5.29 mm, and improves motion-direction similarity from 0.247 to 0.403. At 288p, however, the dedicated model retains better visual quality and motion accuracy. Resolution mixing thus offers a practical route to higher-resolution generation, with a trade-off in lower-resolution performance.

\begin{table}[!htbp]
\centering\footnotesize
\caption{\textbf{Mixed-resolution training.} Results are grouped by output resolution; the mixed model is evaluated at both resolutions.}
\label{tab:h3-resolution}
\setlength{\tabcolsep}{5pt}
\begin{tabular*}{\textwidth}{@{\extracolsep{\fill}}lcccccc@{}}
\toprule
\tablehead[l]{Variant / output} & \tablehead{PSNR$\uparrow$} & \tablehead{SSIM$\uparrow$} & \tablehead{LPIPS$\downarrow$} & \tablehead{VSC$\downarrow$} & \tablehead{Dir.$\uparrow$} & \tablehead{Grip.$\uparrow$}\\
\midrule
480p training / 480p & 16.09 & 0.723 & 0.2598 & 16.11 & 0.247 & 0.994\\
Mixed training / 480p & \textbf{18.20} & \textbf{0.757} & \textbf{0.2036} & \textbf{10.82} & \textbf{0.403} & \textbf{0.995}\\
\midrule
288p training / 288p & \textbf{19.88} & \textbf{0.757} & \textbf{0.1536} & \textbf{9.71} & \textbf{0.443} & \textbf{0.996}\\
Mixed training / 288p & 18.14 & 0.723 & 0.1906 & 13.59 & 0.397 & \textbf{0.996}\\
\bottomrule
\end{tabular*}
\end{table}

\section{Detailed Benchmark Results}
\label{app:benchmark-results}

Tables~\ref{tab:worldarena-full} and~\ref{tab:triworldbench-full} provide the complete results for the retained WorldArena metrics and all 19 TriWorldBench submetrics, respectively. Scores are reported on a 0--100 scale; higher is better. Bold and underlined values indicate the best and second-best results among the compared models.

\paragraph{Frame-rate sensitivity in WorldArena.}
WorldArena's motion-quality metrics are strongly affected by frame rate: lowering the evaluation frame rate can substantially increase these scores without a corresponding improvement in generated video quality. We therefore omit motion quality and the aggregate EWMScore from our reported WorldArena tables, including the main-paper comparison, rather than select a lower frame rate solely to improve benchmark scores.

\par\addvspace{4pt}
\noindent\begin{minipage}{\linewidth}
\makeatletter\def\@captype{table}\makeatother
\centering\footnotesize
\caption{WorldArena Track~1 results on the retained metrics.}
\label{tab:worldarena-full}
\setlength{\tabcolsep}{3pt}
\begin{tabular*}{\textwidth}{@{\extracolsep{\fill}}lcccccc@{}}
\toprule
 & \multicolumn{3}{c}{\tablehead{Visual quality}} & \multicolumn{3}{c}{\tablehead{Content consistency}}\\
\cmidrule(lr){2-4}\cmidrule(l){5-7}
\tablehead[l]{Model} & \tablehead{IQ} & \tablehead{Aes} & \tablehead{JEPA} & \tablehead{Subj} & \tablehead{Bg} & \tablehead{Photo}\\
\midrule
Ctrl-World & 42.44 & 37.05 & 92.77 & \textbf{83.56} & \underline{90.30} & 12.88\\
GigaWorld-1 & 51.18 & \underline{41.17} & 96.77 & 80.97 & 86.43 & \underline{28.11}\\
ABot-PhysWorld & \textbf{61.03} & \textbf{41.83} & 90.36 & 80.56 & 89.02 & 20.52\\
GE-Sim 2.0 & 46.73 & 39.07 & \underline{97.16} & \underline{82.50} & \textbf{90.80} & 7.05\\
Xiaomi-Robotics-U0 & \underline{53.94} & 40.79 & 90.60 & 79.05 & 86.44 & 2.13\\
\ours{} & 45.82 & 39.91 & \textbf{98.39} & 78.91 & 85.70 & \textbf{35.34}\\
\bottomrule
\end{tabular*}
\vspace{4pt}
\begin{tabular*}{\textwidth}{@{\extracolsep{\fill}}lcccccc@{}}
\toprule
 & \multicolumn{2}{c}{\tablehead{Physics adherence}} & \multicolumn{2}{c}{\tablehead{3D accuracy}} & \multicolumn{2}{c}{\tablehead{Controllability}}\\
\cmidrule(lr){2-3}\cmidrule(lr){4-5}\cmidrule(l){6-7}
\tablehead[l]{Model} & \tablehead{Inter} & \tablehead{Traj} & \tablehead{Depth} & \tablehead{Persp} & \tablehead{Instr} & \tablehead{Sem}\\
\midrule
Ctrl-World & 62.62 & 48.20 & 93.25 & 83.66 & 67.68 & 88.68\\
GigaWorld-1 & 75.10 & 54.27 & \textbf{98.44} & 95.60 & 82.14 & 89.42\\
ABot-PhysWorld & \underline{81.96} & 31.50 & 71.99 & \textbf{98.94} & \underline{92.10} & \underline{89.58}\\
GE-Sim 2.0 & 73.60 & \underline{58.63} & 96.49 & 93.64 & 82.70 & 89.05\\
Xiaomi-Robotics-U0 & \textbf{87.30} & 41.89 & 85.25 & \underline{98.84} & \textbf{93.86} & 89.35\\
\ours{} & 72.25 & \textbf{62.19} & \underline{98.36} & 92.85 & 81.66 & \textbf{90.41}\\
\bottomrule
\end{tabular*}
\par\vspace{2pt}\parbox{\textwidth}{\scriptsize Abbreviations: IQ, image quality; Aes, aesthetic quality; JEPA, JEPA similarity; Subj, subject consistency; Bg, background consistency; Photo, photometric consistency; Inter, interaction quality; Traj, trajectory accuracy; Depth, depth accuracy; Persp, perspectivity; Instr, instruction following; Sem, semantic alignment.}
\end{minipage}\par\addvspace{4pt}

\par\addvspace{4pt}
\noindent\begin{minipage}{\linewidth}
\makeatletter\def\@captype{table}\makeatother
\centering\scriptsize
\caption{TriWorldBench results: TWB-Score and all 19 submetrics.}
\label{tab:triworldbench-full}
\setlength{\tabcolsep}{3pt}
\begin{tabular*}{\textwidth}{@{\extracolsep{\fill}}lcccccccccc@{}}
\toprule
 & \tablehead{Score} & \multicolumn{6}{c}{\tablehead{Tri-view consistency}} & \multicolumn{3}{c}{\tablehead{Task alignment}}\\
\cmidrule(lr){3-8}\cmidrule(l){9-11}
\tablehead[l]{Model} & \tablehead{TWB} & \tablehead{N-PSNR} & \tablehead{SSIM} & \tablehead{VLM-I} & \tablehead{VLM-II} & \tablehead{VLM-III} & \tablehead{VQA} & \tablehead{Instr} & \tablehead{Sem} & \tablehead{JEPA}\\
\midrule
DreamDojo & \underline{51.72} & 59.78 & 71.58 & \underline{77.14} & \underline{73.26} & \underline{88.19} & \underline{47.82} & 49.72 & \underline{88.28} & \underline{30.73}\\
Motus & 42.35 & \underline{60.13} & \underline{76.31} & 63.68 & 67.56 & 87.22 & 45.33 & \underline{52.28} & 86.54 & 10.24\\
Genie Envisioner & 40.73 & 58.74 & 74.44 & 69.40 & 57.13 & 77.93 & 36.72 & 20.20 & 79.30 & 0.02\\
Ctrl-World & 38.98 & 46.39 & 64.62 & 46.44 & 63.69 & 82.46 & 40.90 & 38.90 & 81.94 & 10.32\\
\ours{} & \textbf{53.68} & \textbf{62.55} & \textbf{79.06} & \textbf{78.15} & \textbf{80.73} & \textbf{93.32} & \textbf{54.05} & \textbf{62.32} & \textbf{88.75} & \textbf{42.08}\\
\bottomrule
\end{tabular*}
\vspace{4pt}
\begin{tabular*}{\textwidth}{@{\extracolsep{\fill}}lcccccccccc@{}}
\toprule
 & \multicolumn{2}{c}{\tablehead{Physical/3D}} & \multicolumn{3}{c}{\tablehead{Motion quality}} & \multicolumn{3}{c}{\tablehead{Temporal consistency}} & \multicolumn{2}{c}{\tablehead{Visual quality}}\\
\cmidrule(lr){2-3}\cmidrule(lr){4-6}\cmidrule(lr){7-9}\cmidrule(l){10-11}
\tablehead[l]{Model} & \tablehead{Inter} & \tablehead{Persp} & \tablehead{State} & \tablehead{Flow} & \tablehead{Traj} & \tablehead{Subj} & \tablehead{Bg} & \tablehead{Photo} & \tablehead{IQ} & \tablehead{Aes}\\
\midrule
DreamDojo & \textbf{28.28} & 59.39 & \underline{60.76} & 6.52 & \underline{16.61} & 77.37 & \textbf{89.29} & \textbf{15.86} & \underline{24.45} & \underline{17.58}\\
Motus & 25.00 & 44.20 & 59.31 & 0.80 & 13.77 & 77.37 & 1.00 & 1.31 & 16.08 & 16.44\\
Genie Envisioner & 20.10 & \textbf{87.91} & 59.28 & 0.40 & 0.84 & \textbf{91.40} & 1.07 & 4.08 & 19.36 & 15.57\\
Ctrl-World & 24.19 & 45.15 & 60.47 & \textbf{21.29} & 6.09 & 77.05 & \underline{49.02} & \underline{12.44} & 18.37 & 14.92\\
\ours{} & \underline{28.24} & \underline{74.82} & \textbf{62.01} & \underline{17.82} & \textbf{30.39} & \underline{83.86} & 25.45 & 7.16 & \textbf{26.99} & \textbf{22.09}\\
\bottomrule
\end{tabular*}
\par\vspace{2pt}\parbox{\textwidth}{\scriptsize N-PSNR, normalized PSNR; VLM-I/II/III, the three VLM consistency protocols; VQA, VQA consistency; Instr, instruction following; Sem, semantic alignment; Inter, interaction quality; Persp, perspective; State, state alignment; Traj, trajectory accuracy; Subj, subject consistency; Bg, background consistency; Photo, photometric smoothness; IQ, image quality; Aes, aesthetic quality.}
\end{minipage}\par\addvspace{4pt}

\section{Additional Qualitative Observation}
\label{app:view-recovery}

\paragraph{Cross-view recovery.}
We observe that \ours{} can use the other two camera views to recover content in a target view. Figure~\ref{fig:view-recovery} shows qualitative examples alongside the corresponding ground-truth frames. The recovered views capture the broad scene layout and object appearance, although fine details can differ. We report this as an additional qualitative observation of the model's cross-view prediction capability.

\par\addvspace{2pt}
\noindent\begin{minipage}{\linewidth}
\makeatletter\def\@captype{figure}\makeatother
\setlength{\abovecaptionskip}{3pt}
\centering\includegraphics[width=\linewidth]{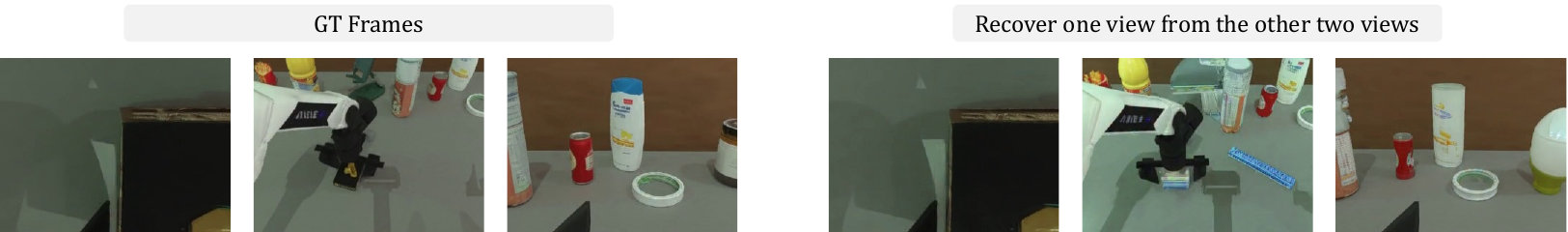}
\caption{\textbf{Cross-view recovery.} Ground-truth frames (left) and target views recovered using the other two camera views (right).}
\label{fig:view-recovery}
\end{minipage}

\end{document}